\documentclass{article}

 \usepackage[dblblindworkshop, final]{neurips_2025}

\usepackage[utf8]{inputenc} 
\usepackage[T1]{fontenc}    
\usepackage{hyperref}       
\usepackage{url}            
\usepackage{booktabs}       
\usepackage{amsfonts}       
\usepackage{nicefrac}       
\usepackage{microtype}      
\usepackage{xcolor}         

\usepackage{graphicx}
\usepackage{amsmath}

\usepackage{lineno}
\usepackage{booktabs}   
\usepackage{tabularx}   
\usepackage{colortbl}   
\usepackage[table]{xcolor}

\usepackage{tcolorbox}
\tcbuselibrary{listings, skins, breakable}
\usepackage{float}
\usepackage{xcolor}
\usepackage{listings}
\usepackage[ruled,vlined,noend]{algorithm2e}

\title{The Conductor and the Engine: A Path Towards Co-Designed Reasoning}

\author{%
  Yuanxin Wang \\
  \texttt{michael.wang@cerebras.net} \\
  \And
  Pawel Filipczuk \\
  \texttt{pawel.filipczuk@cerebras.net} \\
  \AND
  Anisha Garg \\
  \texttt{anisha.garg@cerebras.net} \\
  \And
  Amaan Dhada \\
  \texttt{amaan.dhada@cerebras.net} \\
  \And
  Mohammad Hassanpour \\
  \texttt{mohammad.hassanpour@cerebras.net} \\
  \And
  David Bick \\
  \texttt{david.bick@cerebras.net} \\
  \And
  Ganesh Venkatesh \\
    \texttt{ganesh.venkatesh@cerebras.net} \\
  \\
  \textbf{\textsc{Applied AI Research, Cerebras}}\\
}

\begin{document}
\newcommand{\ignore}[1]{}
\newcommand{\cepo}{\textsc{Coda}}
\newcommand{\ceponc}{\textsc{Coda}\textit{ w/o coding}}
\newcommand{\cepof}{Conductor-driven Architecture (\textsc{Coda})}

\maketitle

\begin{abstract}

Modern LLM reasoning relies on extensive test-time computation, driven by internal model training and external agentic orchestration. However, this synergy is often inefficient, as model verbosity and poor instruction following lead to wasted compute. We analyze this capability-cost trade-off and introduce an optimized reasoning workflow (\cepo) that empowers smaller open-source models to outperform models multiple times their size. We will open-source this workflow to enable further research. Our work demonstrates a clear path toward co-designing orchestration frameworks with the underlying model capabilities to unlock powerful reasoning in small-to-medium sized models.

\end{abstract}

\section{Introduction}

\ignore{
Points to make in the paper
\begin{itemize}
    \item Reasoning at inference time central to the recent boost in LLM's ability to solve extremely challenging problems in math, coding, games etc.
    \item Two emerging trends: i) models being trained with RL to have "internal" reasoning abilities. ii) external prompting and agentic flow to allow models to solve extremely challenging problems by breaking a complex problem down into steps and working through them in an iterative manner
    \item Challenge: effort duplication given both the model and the orchestration attempting to do reasoning. Furthermore, weak instruction following by models can further introduce inefficiencies. 
    \item This paper: i) Analysis of challenges in the test-time orchestration + reasoning model paradigm, ii) Test-time orchestration for enhancing reasoning models ii) Test-time problem reformulation to further enhance model capability -- allowing the models to use the most efficient approach for each problem instead of brute force scaling up model size.
    \item Conclusion: these results provide a good blueprint for how to enhance reasoning models for test-time compute and how co-designing orchestration with reasoning models.
\end{itemize}
}
The ability of Large Language Models (LLMs) to solve exceptionally complex problems in domains like mathematics, software development, and strategic games is increasingly unlocked by substantial computation at inference time~\cite{o1,deepseekr1,qwen,geminideepthink}. Progress in this domain is advancing along two parallel fronts: i) \textbf{Reasoning Engine:} training models with techniques like reinforcement learning to cultivate implicit, "internal" reasoning capabilities~\cite{grpo,ppo}, ii) \textbf{Conductor}: external, agentic workflows that decompose tasks into iterative, verifiable steps~\cite{geminideepthink,mcts}. While the combination of these internal model capabilities and external frameworks holds enormous promise, it also introduces critical new challenges in efficiency and scalability which we evaluate and address in this paper.

\ignore{Challenges}
The first of these challenges is effort duplication. When both the external orchestration framework (the "Conductor") and the internal model (the "Engine") attempt to perform high-level reasoning, they can produce redundant or conflicting operations, leading to wasted compute. This inefficiency is compounded by a second, pervasive challenge: weak instruction following~\cite{lrmif,qwenif}. These powerful reasoning models often struggle to adhere perfectly to complex instructions, causing the agentic flow to deviate, consume excess resources, and ultimately fail. Together, these issues create a significant barrier to deploying robust and capable reasoning systems at scale.

\ignore{our work}
To overcome these obstacles, this paper introduces \cepof, an adaptive framework that actively orchestrates the reasoning Engine. Our approach amplifies model capabilities, enabling medium-sized models to achieve state-of-the-art results that surpass models multiple times their size across challenging tasks in scientific reasoning, math and code generation. This provides a clear and efficient path towards the next generation of powerful AI systems - ones that deliver elite performance and can be deployed on systems with practical memory budgets.
\section{The Conductor: A Framework for State-of-the-Art Reasoning}

\ignore{
Points to make:
\begin{itemize}
    \item Our approach uses a combination of orchestration and problem formulation. 
    \item In terms of orchestration -- planning, self-reflection, cross pollination of ideas from multiple attempts
    \item In terms of problem formulation -- problem formulation to invoke different tools or iterate based on feedback on the current solution
    \item Achieve leaderboard topping results using test-time compute plus two quality reasoning models: Qwen3 32B and GPT-OSS
    \item Rest of the paper goes into the importance of these components and future work on further enhancing their abilities.
\end{itemize}
}
\begin{figure}[!t]
\centering
\includegraphics[width=0.75\columnwidth]{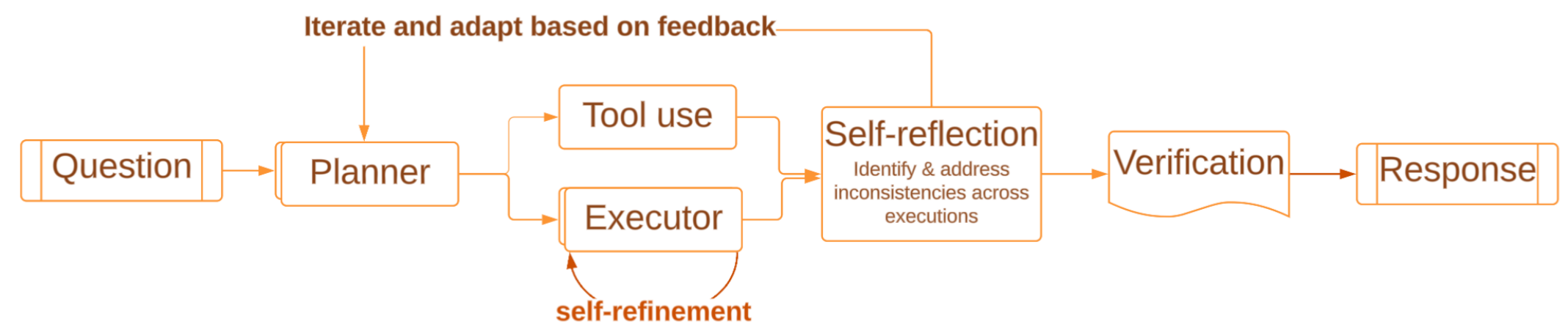}
\caption{\label{fig:approach} \textbf{Overview of \cepo\ Framework}. Simplified view of our orchestration framework}
\vspace{-8pt}
\end{figure}

\cepo\ (simplified view in Figure~\ref{fig:approach}) is composed of several key components that work in concert. It begins with an \textbf{adaptive planner} that assesses a given problem and can map out multiple distinct execution paths to find a solution. The framework then passes these plans to \textbf{executor}, which attempts to solve the problem. A crucial element of executor is \textbf{iterative self-refinement}, where the system can analyze feedback from its own solution attempts such as the output from a code execution and adapt its approach (such as \textbf{problem reformulation}) accordingly. The executor phase produces multiple possible executions which is then passed onto \textbf{self-reflection} phase to synthesize insights from multiple attempts and facilitate  ``cross-pollination'' of ideas by identifying and addressing inconsistencies across different executions. Final step of our orchestration is \textbf{Verification} which can invoke multiple methods such as majority vote or LLM-as-a-judge based on the problem setting.

The above formulation (Algorithm \ref{alg:coda-simple}, Appendix \ref{app:adaptive_flow}) enables us to dynamically allocate more compute to more difficult problems by invoking multiple stages of problem reformulation and self-refinement while the simple queries can pass through without invoking all the stages of the pipeline. This ``right-tool-for-the-job'' approach allows the framework to invoke a variety of external tools as needed including Python interpreter, code compilation and verification. 

The effectiveness of \cepo\ is demonstrated in Table~\ref{tab:ceporesults} and Table~\ref{tab:ceporesultsaa} by its ability to elevate strong open-source models to the top of leaderboards,
often outperforming much larger commercial models. For Qwen~\cite{qwen} family of models, applying our orchestration allows Qwen3 32B to outperform much larger models such as DeepSeek R1, Qwen3 235B and even closed-source models such as OpenAI O3-mini~\cite{o3mini}. To demonstrate the generality of our approach, we also applied it to the recently released GPT-OSS~\cite{gptoss} models in a manner compliant with the Artificial Analysis~\cite{aa} benchmark suite. On various tasks\footnote{GPQA results are averaged across three runs.}, our framework boosts these models to the top tier of leaderboards\footnote{The AIME benchmark was excluded for this model as its baseline performance already approached 95\%.}.

\begin{table*}[!t] 
\caption{
    Our framework, \cepo, elevates the medium-sized Qwen3 model to achieve state-of-the-art (SOTA) performance, outperforming much larger open-source models as well as leading closed-source reasoning systems on math and coding benchmarks. Evaluation setup details in Appendix \ref{app:eval_details}.
    \label{tab:ceporesults}
}
\centering 
\small 
\begin{tabularx}{\textwidth}{l>{\centering\arraybackslash}X>{\columncolor{green!10}\centering\arraybackslash}X>{\centering\arraybackslash}X>{\columncolor{green!10}\centering\arraybackslash}X *{4}{>{\centering\arraybackslash}X}} 
\toprule
\textbf{Benchmark} & \textbf{Qwen3 8B} & \textbf{Qwen3 8B + \cepo} & \textbf{Qwen3 32B} & \textbf{Qwen3 32B + \cepo} & \textbf{Qwen3 235B} & \textbf{DeepSeek R1~\cite{deepseekr1}} & \textbf{o3-mini med.~\cite{o3mini}} & \textbf{Grok3 Think~\cite{grok3}} \\
\midrule
\textsc{AIME 2024 \cite{aime}} & 74.00 & 86.70 & 81.40 & 90.70 & 85.70 & 79.80 & 79.60 & 83.90 \\
\textsc{AIME 2025 \cite{aime}} & 68.33 & 80.00 & 72.90 & 83.30 & 81.50 & 70.00 & 74.80 & 77.30 \\
\textsc{GPQA \cite{gpqa}} & 59.25 & 62.45 & 66.83 & 70.03 & 71.10 & 71.50 & 76.80 & 80.20 \\
\textsc{LiveCodeBench \cite{lcb}} & 55.69 & 60.48 & 65.70 & 71.86 & 70.70 & 64.30 & 66.30 & 70.60 \\

\bottomrule
\end{tabularx}
\vspace{-10pt}
\end{table*}



\begin{table*}[!t] 
\caption{
    \cepo\ turbocharges recent GPT-OSS~\cite{gptoss} models to top positions of Artificial Analysis leaderboard for multiple tasks~\cite{aa}. 
     Evaluation setup details in Appendix \ref{app:eval_details}.
    \label{tab:ceporesultsaa}
}
\centering 
\small 
\begin{tabularx}{\textwidth}{l>{\centering\arraybackslash}X>{\columncolor{green!10}\centering\arraybackslash}X>{\centering\arraybackslash}X>{\columncolor{green!10}\centering\arraybackslash}X *{4}{>{\centering\arraybackslash}X}} 
\toprule
\textbf{Benchmark} & \textbf{gpt-oss-20b} & \textbf{gpt-oss-20b + \cepo} & \textbf{gpt-oss-120b}  & \textbf{gpt-oss-120b + \cepo} & \textbf{Qwen3 235B 2507~\cite{qwen}} & \textbf{Gemini 2.5 Pro~\cite{geminideepthink}} & \textbf{Grok4~\cite{grok4}} \\
\midrule
\textsc{LiveCodeBench~\cite{lcb}} & 72.10 & 82.01 & 76.82 & 87.51 & 79.00 & 80.00 & 82.00 \\
\textsc{SciCode~\cite{scicode}} & 35.40 & 40.10 & 36.20 & 41.00 & 42.00 & 43.00 & 46.00 \\
\textsc{GPQA~\cite{gpqa}} & 70.70 & 76.01 & 76.50 & 82.57 & 79.00 & 84.40 & 87.70 \\

\bottomrule
\end{tabularx}
\vspace{-10pt}
\end{table*}

The rest of the paper will delve deeper into the importance of these individual components, with detailed analysis and ablations. We will conclude with a discussion of future work focused on further enhancing the capabilities of the orchestration via co-design with next-generation reasoning models.

\section{Deconstructing the Conductor: A Component-wise Analysis}
\label{sec:ablations}

\begin{table*}[htbp] 
\caption{
    Performance comparison of Qwen3~\cite{qwen} models, with and without components of our proposed \cepo\ framework. Evaluation configuration details shown at Appendix \ref{app:eval_details}.
    \label{tab:cepoabqwen}
}
\centering 
\small 
\begin{tabularx}{\textwidth}{l *{8}{>{\centering\arraybackslash}X}} 
\toprule
\textbf{Config} & \textbf{AIME 2025} & \textbf{GPQA} & \textbf{LiveCodeBench} & \textbf{SciCode} \\ 
\midrule

Qwen3 8B & 68.33 & 59.25 & 58.05 & 31.70  \\\rowcolor{green!10}
\cepo & 72.66 & 62.45 & 61.58 & 37.10 \\ 
w/o planner  & 74.66 & 59.42 & 60.63 & 37.00 \\
w/o self-reflection & 69.72 & 60.65 & 59.61 & 19.60  \\
recall@best\_of\_N & 79.33 & 71.21 & 64.76 &  40.10\\
\bottomrule
Qwen3 32B & 74.00 & 66.83 & 66.77 & 35.40  \\\rowcolor{green!10}
\cepo & 78.00 & 70.03 & 71.21 &  40.90\\
w/o planner  & 79.33 & 66.83 & 71.21 &  42.80\\
w/o self-reflection & 76.00  & 67.27 & 67.40  &  33.40\\
recall@best\_of\_N & 83.33 & 76.43 & 72.69 &  46.10\\
\bottomrule

\end{tabularx}
\vspace{-10pt}
\end{table*}

\begin{table*}[htbp] 
\caption{
    Performance of GPT-OSS~\cite{gptoss} benefits from all the components of our \cepo\ framework while showing room for accuracy boost by improving model capabilities in planning and verification.
    \label{tab:cepoabgptoss}
}
\centering 
\small 
\begin{tabularx}{\textwidth}{l *{8}{>{\centering\arraybackslash}X}} 
\toprule
\textbf{Config}  & \textbf{GPQA} & \textbf{LiveCodeBench} & \textbf{SciCode} \\ 
\midrule

gpt-oss-20b &  70.70 & 72.10 & 35.40  \\\rowcolor{green!10}
\cepo &  76.01 & 82.01 & 40.10 \\
w/o planner  &  72.89 & 81.48 & 42.90 \\
w/o self-reflection &  73.57 & 77.40 & 38.20 \\
recall@best\_of\_N &  81.31 &  83.17 &   45.50\\
\bottomrule
gpt-oss-120b  & 76.50 & 76.82 & 36.20  \\\rowcolor{green!10}
\cepo  & 82.57 & 87.51 & 41.00 \\
w/o planner  &  76.76 & 87.19 & 41.00\\
w/o self-reflection  & 78.84 & 82.79 &  38.80\\
recall@best\_of\_N  & 87.62 & 88.25 &  44.10\\
\bottomrule

\end{tabularx}
\vspace{-10pt}
\end{table*}

We systematically evaluate the impact of the core components $-$ planning, self-reflection and cross-solution verification $-$ on a challenging reasoning benchmarks. The results, detailed in Table~\ref{tab:cepoabqwen} and Table~\ref{tab:cepoabgptoss}, reveal the importance of this multi-component orchestration in \cepo\ for achieving high performance on diverse reasoning tasks. Our analysis begins to uncover the critical interplay between the \cepo\ strategy and the nature of the task, highlighting the Conductor's role in maximally leveraging the reasoning Engine to solve challenging problems.

We explore this dynamic by analyzing the framework's impact on different model families and reasoning tasks, which reveals several key observations:

\paragraph{Planning}
The utility of the planning component appears to be highly task-dependent. We observe significant benefits on scientific reasoning benchmarks such as GPQA. However, its direct impact on improving coding performance was more muted in our experiments. This suggests that a reasoning model specifically post-trained to utilize an explicit planning phase could more fully exploit this component.

\paragraph{Self-Reflection}
The iterative self-reflection component provides a consistent and strong accuracy advantage across the models and tasks we evaluated. This benefit likely stems from two sources: the model's intrinsic ability to critique and refine its own work, and the framework's process of synthesizing insights from multiple execution attempts, effectively cross-pollinating ideas to produce a superior final output.

\paragraph{Verification}
The verification step presents a significant opportunity for future improvement. We observe a notable gap between the final performance of our Conductor and the theoretical maximum achievable (i.e., the recall at best-of-N across multiple attempts), indicating that a more accurate verifier could unlock substantial gains. This points to a promising direction for co-design: explicitly training the reasoning Engine to act as a verifier, for instance through reinforcement learning, to more effectively guide the test-time computation.

\section{Dynamic Problem Reformulation for Efficient Reasoning}
\label{sec:reformulation}
\cepo\ employs dynamic problem reformulation, adapting its strategy based on the problem's complexity and the specific strengths of the Engine model. This allows the system to select the most efficient and reliable path to a solution. We illustrate this capability with two examples below.

\paragraph{Adaptive Path Selection} We observe that models have strong implicit preferences for solving problems in a certain way (Table~\ref{tab:preference}); for instance, some mathematical problems are solved more reliably through direct textual reasoning (``mental math''), while others are better suited for code generation and execution. Our orchestration flow (Algorithm \ref{alg:coda-adaptive}, Appendix \ref{app:adaptive_flow}) captures this insight by exploring multiple solution paths and prioritize the one that is most likely to succeed for that specific model. This adaptive approach yields solid gains, boosting the performance of both the Qwen3 8B and Qwen3 32B models on the AIME 2024 and AIME 2025 math benchmarks as shown in Table~\ref{tab:formulation}, beating out larger, math-specific models~\cite{rstar2agent}. More qualitative examples are at Appendix \ref{app:adaptive_examples}.

\begin{table}[htbp]
\centering
\caption{\textbf{Adaptive Path Selection Maximizes Performance} The Pass@10 success rate is highest when the system can choose the optimal solution path, as direct reasoning (``mental math'') and code generation excel on different subsets of problems. Dataset here is 15 hard questions from Numina Math dataset~\cite{numina}.}
\label{tab:preference}
\begin{tabular}{lcc}
\toprule
\textbf{Mental Math only} & \textbf{Coding only} & \textbf{Mental Math or Code} \\
\midrule
93\% & 93\% & 100\%   \\
\bottomrule
\end{tabular}
\vspace{-5pt}
\end{table}

\begin{table}[htbp]
\centering
\caption{Adaptive path selection via code generation provides a significant performance boost for mathematical reasoning. The table shows ablation results where adding the adaptive coding reformulation to \cepo\ yields large gains on the AIME benchmarks.}
\label{tab:formulation}
\resizebox{\textwidth}{!}{%
\begin{tabularx}{1.05\textwidth}{lcc||lcc}
\toprule
\textbf{Model} & \textbf{AIME 2024} & \textbf{AIME 2025} & \textbf{Model} & \textbf{AIME 2024} & \textbf{AIME 2025} \\
\hline
Qwen3 8B & 74.00 & 68.33   & Qwen3 32B & 81.40 & 74.00   \\\rowcolor{red!10}
\ceponc & 83.33 & 72.66   & \ceponc  & 84.00 & 78.00   \\\rowcolor{green!10}
\cepo\ & 86.67 & 80.00   & \cepo & 90.70 & 83.30   \\
\bottomrule
\end{tabularx}
}
\vspace{-5pt}
\end{table}

 \paragraph{Iterative Refinement with Rich Feedback} In complex tasks like code generation, a model's first attempt may not be perfect. When our framework executes a generated piece of code, it captures the execution feedback (e.g., interpreter error messages or incorrect output). The problem is then reformulated and presented back to the model, including the original prompt along with this rich feedback and an instruction to correct its previous mistakes. This iterative refinement loop, where the task is progressively clarified based on execution results, leads to significant improvements in code generation accuracy (Table~\ref{tab:feedback}).

\begin{table}[htbp]
\centering
\caption{Performance gains for the Qwen3 8B and Qwen3 32B models on the LiveCodeBench benchmark using iterative self-refinement. Evaluation configuration details shown at Appendix ~\ref{app:eval_details}.}
\label{tab:feedback}
\begin{tabularx}{0.6\textwidth}{lcc>{\columncolor{green!10}\centering\arraybackslash}X}
\toprule
\textbf{Model} & \textbf{Baseline} & \textbf{\cepo} & \textbf{\cepo\ with Tests} \\
\midrule
Qwen3 8B  & 55.69 & 56.29 & 60.48   \\
Qwen3 32B  & 65.70 & 67.07 & 71.86   \\
\bottomrule
\end{tabularx}
\vspace{-5pt}
\end{table}

\ignore{
\begin{itemize}
    \item Treat problems that are best solved using different approach appropriately. We give two examples of reformulation in this section.
    \item For example: Qwen3 8B prefers to solve some mathematical problems directly, some through coding and for some either approach works
    \item Updating the orchestration flow to capture this insight shows solid gains for  Qwen3 8B and Qwen3 32B on AIME 2024 and AIME 2025
    \item Second example is refining the task by giving rich feedback -- retrying code generation with execution feedback. Results show gains for Qwen3 8B and Qwen3 32B on LiveCodeBench.
\end{itemize}
}

\section{Conclusion}
\label{sec:conclude}

In this work, we demonstrated that intelligent orchestration is a parameter-efficient path to elite performance. Our framework propelled medium-sized open-source models, such as the Qwen3 32B and GPT-OSS, to the top of competitive leaderboards, allowing them to outperform models multiple times their size. These findings point towards a future of co-designing LLMs and their reasoning frameworks. Instead of relying on rigid, fixed templates, orchestration should be dynamically guided by model capabilities and problem complexity. The next frontier is to move beyond monolithic reasoning engines and forge models with a toolkit of capabilities - such as planning, task decomposition, verification, reflection - enabling the conductor to transform from a mere prompter into a true director of computational thought.
\bibliographystyle{plain}   
\bibliography{references}   

\appendix
\begingroup
\newcommand{\divider}{\par\noindent\rule{\linewidth}{0.4pt}\par}
\section{Related Work}
\label{app:related_work}

Many prior work on reasoning models has largely focused on instilling advanced thinking capabilities into model weights through multiple stages of training — from pretraining \cite{readquiet, rho1}, supervised fine-tuning \cite{ye2025limo, muennighoff2025s1}, to pruning \cite{sui2025stop} and reinforcement learning \cite{ppo, grpo, gspo, fatemi2025concise, deepscaler2025}. 

In contrast to these training-based approaches, an emerging line of research investigates training-free algorithms that operate purely at inference time. State-of-the-art reasoning systems such as OpenAI’s o1 \cite{o1} and o3-mini \cite{o3mini}, DeepSeek R1 \cite{deepseekr1}, and Gemini 2.5 \cite{geminideepthink} achieve their performance largely by allocating substantial computation at inference. Techniques like Chain-of-Thought prompting \cite{cot}, Monte Carlo Tree Search \cite{mcts}, and CoT-decoding \cite{cotdecoding} improve reasoning by broadening exploration of intermediate thinking trajectories and systematically breaking down complex problems into clear sub-tasks. 

Foundational test-time strategies, such as Chain-of-Thought and Monte Carlo Tree Search, emerged in the era of large-scale LLMs and were often designed to leverage their massive parameter counts, creating a high barrier to entry. However, the recent advent of powerful, parameter-efficient reasoning models challenges this dependency on scale and warrants new research into orchestration techniques built to maximize their unique capabilities. Therefore, recent research has moved toward building multi-agent, iterative frameworks on top of \textbf{smaller LLMs} \cite{cerebras2024ceporeasoning, cerebras2025cepoupdate, wang2024mixture}, with support of ultra long-context, \cite{longcontextrag, cerebras2025longcepo}, and integrating planning, decomposition, self-reflection, and verification into unified orchestration pipelines.

Our work builds on this trajectory by demonstrating that carefully designed test-time orchestration can amplify small- and medium-scale models to outperform models many times their size, offering a practical path toward scalable and cost-efficient reasoning systems.
\section{Evaluations Details}
\label{app:eval_details}
For Qwen3 family results in Table ~\ref{tab:ceporesults} and ~\ref{tab:feedback}, we use the evaluation setup in Qwen3 Technical Report \cite{qwen} and reach parity with the reported scores for both Qwen3-8B and Qwen3-32B models. In our orchestration framework, for AIME, we adopt our adaptive path selection algorithm which allows the model to solve the problem using code; for LiveCodeBench, we utilize public tests and execution feedback for iterative self-refinement, with the same date range and release version used in \cite{qwen}: release v5 from 2024-08-01 to 2025-02-01.

For gpt-oss family results in Table ~\ref{tab:ceporesultsaa} and all component-wise analysis in Table ~\ref{tab:cepoabqwen} and ~\ref{tab:cepoabgptoss}, we run the benchmarks in Artificial Analysis \cite{aa} compliant way and reach parity for both gpt-oss-20B and gpt-oss-120B models. For LiveCodeBench, we didn't use public tests and execution feedback for LiveCodeBench since this is not used by Artificial Analysis; and use the same date range and release version in \cite{aa}: release v6 from 2024-07-01 to 2025-01-01.
\section{CODA and Adaptive Path Selection Algorithmic Illustration}\label{app:adaptive_flow}
\begin{algorithm}[H]
\caption{CODA-Simple (w/o Adaptive Path Selection)}
\label{alg:coda-simple}
\DontPrintSemicolon
\SetKwInOut{Input}{Input}
\SetKwInOut{Output}{Output}
\SetKwInOut{Parameter}{Parameters}

\Input{
  Question $\mathcal{Q}$ \\
  Planning Agent $\theta_{\text{plan}}$ \\
  Execution Agent $\theta_{\text{execute}}$ \\
  Self-Reflection Agent $\theta_{\text{reflection}}$ \\
  Verification Agent $\theta_{\text{verify}}$
}
\Parameter{
  $\texttt{best\_of\_n}$: number of attempts \\
  $\texttt{n\_plans}$: number of plans and executions per attempt
}
\Output{
  Final answer $\hat{A}$
}

$\mathcal{S} \gets [\ ]$ \;

\For{$i \gets 1$ \KwTo $\texttt{best\_of\_n}$}{
    $\mathcal{E} \gets [\ ]$ \;

    \For{$j \gets 1$ \KwTo $\texttt{n\_plans}$}{
        $\pi_j \gets \theta_{\text{plan}}\!\left(\mathcal{Q}\right)$ \;

        $e_j \gets \theta_{\text{execute}}\!\left(\mathcal{Q}, \pi_j\right)$ \tcp*{optionally apply self-refinement here} 
        $\mathcal{E} \gets \mathcal{E} \cup \{e_j\}$ \;
    }

    $s_i \gets \theta_{\text{reflection}}\!\left(\mathcal{E}\right)$ \;
    $\mathcal{S} \gets \mathcal{S} \cup \{s_i\}$ \;
}

$\hat{A} \gets \theta_{\text{verify}}\!\left(\mathcal{Q}, \mathcal{S}\right)$ \;

\Return $\hat{A}$ \;

\end{algorithm}


\begin{algorithm}[h]
\caption{CODA-Adaptive for Math Problems}
\label{alg:coda-adaptive}
\DontPrintSemicolon
\SetKwInOut{Input}{Input}
\SetKwInOut{Output}{Output}
\SetKwInOut{Parameter}{Parameters}

\Input{
  Question $\mathcal{Q}$ \\
  Planning Agent $\theta_{\text{plan}}$ \\
  Execution Agent $\theta_{\text{execute}}$ \\
  Self-Refinement Agent $\theta_{\text{reflection}}$ \\
  Verification Agent $\theta_{\text{verify}}$ \\
  Coding Agent $\theta_{\text{code}}$
}
\Parameter{
  $\texttt{num\_attempts\_coda}$: number of attempts to use CODA-Simple flow \\
  $\texttt{num\_attempts\_baseline}$: number of attempts to use the executor agent directly \\
  $\texttt{num\_attempts\_coding}$: number of attempts to use the coding agent \\
  $\texttt{n\_plans}$: number of plans and executions per attempt inside CODA-Simple \\
  $\texttt{best\_of\_n}$: number of attempts inside CODA-Simple \\
}
\Output{
  Final answer $\hat{A}$
}

\BlankLine
\textbf{Helper routines:}\;
\Indp
\texttt{StrictMajority}$(\mathcal{A})$: returns $(\text{has\_maj}, a^\star)$ where $a^\star$ occurs $>\frac{|\mathcal{A}|}{2}$ times; otherwise $(\text{false}, \bot)$.\;
\texttt{PluralityVote}$(\mathcal{A})$: returns the mode(s) $\mathcal{M}$ with maximal frequency (may contain ties).\;
\Indm
\BlankLine

$\mathcal{S} \gets [\ ]$ \tcp*{collector for all candidate answers}

\textbf{Baseline (direct executor)}\;
$\mathcal{B} \gets [\ ]$\;
\For{$i \gets 1$ \KwTo $\texttt{num\_attempts\_baseline}$}{
    $b_i \gets \theta_{\text{execute}}\!\left(\mathcal{Q}\right)$\;
    $\mathcal{B} \gets \mathcal{B} \cup \{b_i\}$\;
}
$\mathcal{S} \gets \mathcal{S} \cup \mathcal{B}$\;

\If{$|\texttt{PluralityVote}(\mathcal{B})| = 1$ \textbf{and} $\texttt{StrictMajority}(\mathcal{B}).\text{has\_maj} = \text{true}$}{
    \Return the unique $b^\star$ from \texttt{PluralityVote}$(\mathcal{B})$\;
}

\BlankLine
\textbf{Coding agent pass}\;
$\mathcal{C} \gets [\ ]$\;
\For{$i \gets 1$ \KwTo $\texttt{num\_attempts\_coding}$}{
    $c_i \gets \theta_{\text{code}}\!\left(\mathcal{Q}\right)$\;
    $\mathcal{C} \gets \mathcal{C} \cup \{c_i\}$\;
}
$\mathcal{S} \gets \mathcal{S} \cup \mathcal{C}$\;

\If{$\texttt{StrictMajority}(\mathcal{C}).\text{has\_maj} = \text{true}$}{
    \Return $\texttt{StrictMajority}(\mathcal{C}).a^\star$\;
}

\BlankLine
\textbf{CODA-Simple passes}\;
$\mathcal{D} \gets [\ ]$ \tcp*{answers from CODA-Simple}
\For{$i \gets 1$ \KwTo $\texttt{num\_attempts\_coda}$}{
    $d_i \gets \text{CODA-Simple}\!\left(\mathcal{Q}, \theta_{\text{plan}}, \theta_{\text{execute}}, \theta_{\text{reflection}}, \theta_{\text{verify}};~\texttt{best\_of\_n};~\texttt{n\_plans}\right)$\;
    $\mathcal{D} \gets \mathcal{D} \cup \{d_i\}$\;
}
$\mathcal{S} \gets \mathcal{S} \cup \mathcal{D}$\;

\BlankLine
\textbf{Global aggregation}\;
$\mathcal{M} \gets \texttt{PluralityVote}(\mathcal{S})$\;
\Return the modes $a^\star \in \mathcal{M}$\;

\end{algorithm}
\section{Examples for Adaptive Path Selection}
\label{app:adaptive_examples}

\definecolor{codebg}{gray}{0.96}
\definecolor{codekw}{RGB}{33,105,175}
\definecolor{codestring}{RGB}{196,85,0}
\definecolor{codecomment}{RGB}{90,110,120}

\lstset{
  language=Python,
  basicstyle=\ttfamily\small,
  backgroundcolor=\color{codebg},
  keywordstyle=\color{codekw}\bfseries,
  stringstyle=\color{codestring},
  commentstyle=\itshape\color{codecomment},
  showstringspaces=false,
  breaklines=true,
  frame=single,
  rulecolor=\color{black!30},
  frameround=tttt
}

\tcbset{
  myexamplebox/.style={
    colback=white,
    colframe=black,
    fonttitle=\bfseries,
    coltitle=white,
    colbacktitle=black!80,
    enhanced,
    attach boxed title to top center={yshift=-2mm},
    boxed title style={colframe=black, colback=black!80, sharp corners},
  }
}
\begin{figure}[H]
\centering
\begin{tcolorbox}[myexamplebox, title={Example 1: AIME 2024 question where CODA w/o code fails for Qwen3-8B}]

\textbf{Question:} \\
Each vertex of a regular octagon is independently colored either red or blue with equal probability. The probability that the octagon can then be rotated so that all of the blue vertices end up at positions where there were originally red vertices is $\tfrac{m}{n}$, where $m$ and $n$ are relatively prime positive integers. What is $m+n$?
\divider
\textbf{Ground Truth Answer:} \\
371
\divider

\textbf{Output from Qwen3-8B + CODA w/o coding (Truncated due to space):} \\
To determine the probability that a randomly colored octagon (with each vertex independently colored red or blue with equal probability) can be rotated so that all blue vertices end up on positions that were originally red, we analyze the combinatorial structure of the problem. \\
**Understanding the Problem**\\
We are given an octagon with 8 vertices, each independently colored ......\\
**Strategy and Key Observations**\\
Let define $ A_k $ as the set of all colorings where \$ B + k, cap B = emptyset,  ......\\
**Final Answer**\\
The probability that the octagon can be rotated so that all blue vertices end up on red positions is $ \tfrac{121}{256} $, where $ m = 121 $ and $ n = 256 $ are relatively prime.Thus, the final result is: $$\boxed{377}$$

\divider
\textbf{Output from Qwen3-8B + CODA with coding:} \\
\begin{lstlisting}[language=Python]
import math
count = 0
for num in range(256):
    color = [(num >> i) & 1 for i in range(8)]
    blue = [i for i in range(8) if color[i] == 1]
    found = False
    for k in range(8):
        valid = True
        for b in blue:
            pos = (b + k) % 8
            if color[pos] != 0:
                valid = False
                break
        if valid:
            found = True
            break
    if found:
        count += 1
g = math.gcd(count, 256)
m = count // g
n = 256 // g
print(m + n)
\end{lstlisting}
\textbf{Code Execution Result:} 371
\divider
This is a geometry problem that requires complex reasoning and very long chain-of-thought, for which Qwen3-8B might struggle. However, the small-size model is able to solve this problem using just a few lines of code.

\end{tcolorbox}
\caption{Example of AIME 2024 question where "mental math" prompt fails but code generation succeeds}
\label{fig:math_vs_coding_aime2024_8b}
\end{figure}


\begin{figure}[p]
\centering
\begin{tcolorbox}[myexamplebox, title={Example 2: AIME 2025 question where CODA w/o code fails for Qwen3-8B}]

\textbf{Question:} \\
Let $N$ denote the number of ordered triples of positive integers $(a,b,c)$ such that $a,b,c\leq3^6$ and $a^3+b^3+c^3$ is a multiple of $3^7$. Find the remainder when $N$ is divided by $1000$.
\divider
\textbf{Ground Truth Answer:} \\
735
\divider

\textbf{Output from Qwen3-8B + CODA w/o coding (Truncated due to space):} \\
To solve the problem, we are asked to count the number of ordered triples $(a, b, c)$ of positive integers such that:- $1 \leq a, b, c \leq 3^6 = 729$- $a^3 + b^3 + c^3$ is divisible by $3^7 = 2187$. Then, we are to compute this number modulo $1000$.

Step 1: Total Number of Triples.Each of $a, b, c$ can independently range from 1 to $729 = 3^6$, so the total number of ordered triples is... $$N_{\text{total}} = (3^6)^3 = 3^{18}$$

Step 2: Modulo Condition. We are interested in the number of such triples $(a, b, c)$... $$a^3 + b^3 + c^3 \equiv 0 \pmod{3^7}$$ 

Step 3: Uniform Distribution AssumptionA key insight is that **modulo $3^k$**, the function $x \mapsto x^3$ is **not uniformly distributed**, as shown in small cases (e.g., modulo 9, where cube residues are not all distinct or evenly distributed)...

Step 4: Final Computation: Now, compute:$$3^{11} = 177147$$Now take modulo $1000$:$$177147\mod 1000 = 147$$

\#\#\# Final Answer$$\boxed{147}$$

\divider
\textbf{Output from Qwen3-8B + CODA with coding:} \\
\begin{lstlisting}[language=Python]
m = 3**7
max_a = 3**6
freq = [0] * m
for a in range(1, max_a + 1):
    r = pow(a, 3, m)
    freq[r] += 1
C = [0] * m
for r1 in range(m):
    for r2 in range(m):
        s = (r1 + r2) % m
        C[s] += freq[r1] * freq[r2]
total = 0
for s in range(m):
    target = (-s) % m
    total += C[s] * freq[target]
print(total % 1000)
\end{lstlisting}
\textbf{Code Execution Result:} 735
\divider
This is a hard combinatorial number theory problem which requires a lot of sequential reasoning and memory, and it is very difficult to fit all these thinking into a 40960 context window of Qwen-3 Model. However, when we ask the model to solve it with code, it only takes a few tokens to reach the correct result.
\end{tcolorbox}
\caption{Example of AIME 2025 question where "mental math" prompt fails but code generation succeeds}
\label{fig:math_vs_coding_aime2025_8b}
\end{figure}


\begin{figure}[p]
\centering
\begin{tcolorbox}[myexamplebox, title={Example 3: AIME 2024 question where CODA w/o code fails for Qwen3-32B}]

\textbf{Question:} \\
Define $f(x)=|| x|-\tfrac{1}{2}|$ and $g(x)=|| x|-\tfrac{1}{4}|$. Find number of intersections of the graphs of \[y=4 g(f(\sin (2 \pi x))) \quad\text{ and }\quad x=4 g(f(\cos (3 \pi y))).\]
\divider
\textbf{Ground Truth Answer:} \\
385
\divider

\textbf{Output from Qwen3-32B + CODA w/o coding (Truncated due to space):} \\
To determine the number of intersections between the graphs of the functions  ...  we begin by analyzing the structure and behavior of the component functions involved.

**Step 1: Understanding the Base Functions...This function is piecewise linear with a \"W\" shape over the interval $[-1, 1]$. It has peaks at $ x = 0 $ and $ x = \pm 1 $ (value $ \tfrac{1}{2} $) and valleys at ... 

**Step 2: Mapping the Composites**\#\#\#\# **Function 1: $ y = 4g(f(\sin(2\pi x))) $** The inner function has a period of 1 and oscillates between $-1$ and $1$. Taking the absolute value gives $ |\sin(2\pi x)| $, which oscillates between ...

**Step 3: Counting Valid Intersections**We now consider the number of **valid intersections** between the graphs. The graphs are defined by:- The first function has **16 segments** with slopes... Thus, the total number of valid intersections is: 96 + 96 = \boxed{192}

\divider
\textbf{Output from Qwen3-32B + CODA with coding:} \\
\begin{lstlisting}[language=Python]
import math
from scipy.optimize import root
import numpy as np
def h(z):
    fz = abs(abs(z) - 0.5)
    return 4 * abs(fz - 0.25)
def equations(vars):
    x, y = vars
    eq1 = y - h(math.sin(2 * math.pi * x))
    eq2 = x - h(math.cos(3 * math.pi * y))
    return [eq1, eq2]
x_values = np.linspace(0, 1, 100)
y_values = np.linspace(0, 1, 100)
solutions = set()
for x0 in x_values:
    for y0 in y_values:
        sol = root(equations, [x0, y0], method='hybr')
        x_sol, y_sol = sol.x
        if 0 <= x_sol <= 1 and 0 <= y_sol <= 1:
        rounded_sol = (round(x_sol, 6), round(y_sol, 6))
        y_check = h(math.sin(2 * math.pi * x_sol))
        x_check = h(math.cos(3 * math.pi * y_sol))
        if abs(y_check - y_sol) < 1e-5 \
        and abs(x_check - x_sol) < 1e-5:
        solutions.add(rounded_sol)
print(len(solutions))
\end{lstlisting}
\textbf{Code Execution Result: 385} 
\divider
This is a complex problem which mixes analysis of piecewise functions, trigonometric composition, and geometric interpretation. Even a medium size Qwen3-32B model struggles to get it correct using CODA without code. Code generation takes the model out of the local optimal in an innovative way.
\end{tcolorbox}
\caption{Another Example of AIME 2024 question where "mental math" prompt fails but code generation succeeds}
\label{fig:math_vs_coding_aime2025_8b}
\end{figure}


\begin{figure}[p]
\centering
\begin{tcolorbox}[myexamplebox, title={Example 4: AIME 2025 question where CODA w/o code fails for Qwen3-32B}]

\textbf{Question:} \\
Alex divides a disk into four quadrants with two perpendicular diameters intersecting at the center of the disk. He draws 25 more line segments through the disk, drawing each segment by selecting two points at random on the perimeter of the disk in different quadrants and connecting those two points. Find the expected number of regions into which these 27 line segments divide the disk.
\divider
\textbf{Ground Truth Answer:} \\
204
\divider

\textbf{Output from Qwen3-32B + CODA w/o coding (Truncated due to space):} \\
Final Review and Error Analysis: The initial solution incorrectly applied the general formula for the expected number of regions formed by random chords:$$E(R) = 1 + \binom{n}{2} + \binom{n}{4}$$ This formula assumes **no constraints** on the endpoints of the chords. However, in the problem, the chords are **constrained** to ...\ Correct Approach: To compute the expected number of regions formed by 27 line segments (2 fixed diameters and 25 random chords connecting points in different quadrants), we proceed in steps: 1. **Initial Setup**- The disk is first divided into 4 quadrants by two perpendicular diameters. These diameters intersect at the center, forming **4 initial regions**.- Alex then draws **25 additional chords**, where each chord connects two points on the circle perimeter in **different quadrants**. The total number of line segments is therefore **27**. 2. **Expected Intersections** ...

Final Answer $$\boxed{\dfrac{287}{3}}$$ This is the **expected number of regions** into which the 27 line segments divide the disk.

\divider

\textbf{Output from Qwen3-32B + CODA with coding:} \\
\begin{lstlisting}[language=Python]
import fractions
# Calculate each part of the expected intersections
I_1 = 1  # Intersection between the two original diameters
I_2 = 25 * 2 * fractions.Fraction(2, 3)  # Intersections between original diameters and added chords
# Number of pairs of added chords
num_pairs_added = (25 * 24) // 2
I_3 = num_pairs_added * fractions.Fraction(17, 36)  # Intersections between pairs of added chords
# Total expected intersections
E_I = I_1 + I_2 + I_3
E_R = 1 + 27 + E_I
print(int(E_R))

\end{lstlisting}
\textbf{Code Execution Result:} 204 
\divider
This is a geometric probability / combinatorial geometry problem. As shown above, CODA without coding struggles to get the correct answer even after extensive self-reflection. In the code generation output, the model calculates some magic numbers in <think> section (too long to fit in) and finish the problem utilizing those numbers precomputed.
\end{tcolorbox}
\caption{Another Example of AIME 2025 question where "mental math" prompt fails but code generation succeeds}
\label{fig:math_vs_coding_aime2025_8b}
\end{figure}

\endgroup


\end{document}